\documentclass[conference]{IEEEtran}
\IEEEoverridecommandlockouts
\usepackage{cite}
\usepackage{amsmath,amssymb,amsfonts}
\usepackage{algorithmic}
\usepackage{graphicx}
\usepackage{textcomp}
\usepackage{xcolor}
\usepackage[T1]{fontenc}
\usepackage[utf8]{inputenc}
\def\BibTeX{{\rm B\kern-.05em{\sc i\kern-.025em b}\kern-.08em
    T\kern-.1667em\lower.7ex\hbox{E}\kern-.125emX}}
\IEEEoverridecommandlockouts\IEEEpubid{\makebox[\columnwidth]{978-1-6654-
9653-7/22/\$31.00~\copyright~2022 IEEE \hfill}
\hspace{\columnsep}\makebox[\columnwidth]{ }}
\begin{document}

\title{Multimodal E-Commerce Product Classification Using Hierarchical Fusion}

\author{\IEEEauthorblockN{Tsegaye Misikir Tashu\IEEEauthorrefmark{1} \IEEEauthorrefmark{3},
Sara Fattouh\IEEEauthorrefmark{1},
P\'eter Kiss\IEEEauthorrefmark{1} and
Tom\'a\v{s} Horv\'ath\IEEEauthorrefmark{1}\IEEEauthorrefmark{2}}
\IEEEauthorblockA{\IEEEauthorrefmark{1} Department of Data Science and Engineering, Faculty of Informatics,
                         ELTE - E\"otv\"os Lor\'and University\\
                        P\'azm\'any P\'eter s\'et\'any 1/C, 1117 Budapest, Hungary }
\IEEEauthorblockA{\IEEEauthorrefmark{2}Faculty of Science, Institute of Computer Science, Pavol Jozef \v{S}af\'arik University, \\ Jesenn\'a 5, 040 01 Ko\v{s}ice, Slovakia}
\IEEEauthorblockA{\IEEEauthorrefmark{3}College of Informatics, Kombolcha Institute of Technology, Wollo University, \\ 208 Kombolcha, Ethiopia}

Email: \IEEEauthorrefmark{1}misikir@inf.elte.hu,
\IEEEauthorrefmark{1}sara.informatics@gmail.com,
\IEEEauthorrefmark{1}peter.kiss@inf.elte.hu,
\IEEEauthorrefmark{1}tomas.horvath@inf.elte.hu
}

\maketitle
\begin{abstract}

In this work, we present a multi-modal model for commercial product classification, that combines features  extracted by multiple neural network models from textual (CamemBERT and FlauBERT) and visual data (SE-ResNeXt-50), using simple fusion techniques.  The proposed method significantly outperformed the performance of the unimodal models, as well as the reported performance of similar models on our specific task. We made experiments with multiple fusing techniques, and found, that the best preforming technique to combine the individual embedding of the unimodal network is based on the combination of concatenation and averaging the feature vectors. Each modality complemented the shortcomings of the other modalities, demonstrating that increasing the number of modalities can be an effective method for improving the performance of multi-label and multimodal classification problems. 
\end{abstract}


\section{Introduction}

Product profiles on e-commerce platforms typically contain unstructured text, product titles, and images, and provide details about the product that is important to e-commerce. The profiles help customers make purchasing decisions and, on the other hand, provide input for automated services of e-commerce sites, such as question-answering and product recommendation systems. Since the number of products and the number of product categories in an e-commerce platform can be so high that it is unmanageable for humans, it is important to have methods that are able to accurately assign these products to their categories \cite{chen-etal-2021}. 

Most research on product classification has focused on the use of text-based cues, even though products on e-commerce platforms contain more than textual information and ignore the valuable information that associated images contain. Therefore, it might be useful to experiment with multimodal models that can combine features extracted from different modalities.

In this work, we proposed a multimodal product classification model based on several pre-trained transformer-based architectures for language modeling and image classification. The transformer-based language models were used to learn a representation of text features from the textual information, and pre-trained visual approaches are used to obtain a representation from images. We used hierarchical fusion to combine the representations obtained from the language models and the visual model, and a fully connected layer to perform the final classification. 

We empirically evaluated the proposed method in the \textit{Multimodal Product Classification Task} of the SIGIR \footnote{https://sigir-ecom.github.io/data-task.html} 2020 E-Commerce Workshop Data Challenge. The results of the experiments show superior performance compared to unimodal and other multimodal methods. Our experiments have also shown that text-based classification models generally perform better than visual models. On the other hand, the inclusion of both modalities allows us to benefit from complimentary feature information and outperformed baseline approaches. \\

The remainder of the paper is organized into six sections. Section~\ref{related} describes related work relevant to our research. Section~\ref{prop} presents our proposed multimodal hierarchical fusion model architecture, section~\ref{experimental} presents the general experimental settings and implementation, while in section~\ref{res} we summarize the experimental results and discussion based on the experiments. Finally, section~\ref{con} presents our conclusions.

\section{Related Work}
\label{related}

In 2020, a large-scale competition to classify multimodal product data was organized as part of the SIGIR 2020 e-commerce workshop. The dataset was provided by Rakuten France, and each product included a title and a detailed description in French, as well as a product image. The task was to predict 27 category labels from four major genres: books, children, households, and entertainment. Several teams and researchers participated \cite{lee2020cbb,bi2020multimodal,chordia2020large,rychalska2020synerise}. The most common solution proposed by most authors was to fine-tune pre-trained text and image models as feature extractors and then use a bimodal fusion mechanism to combine predictions. Most teams used the Transformer-based model BERT \cite{devlin2019bert} for text feature extraction and ResNet \cite{ResNet} for image feature extraction. For bimodal fusion, the methods used were even more diverse. Roughly in order of increasing complexity, the methods included simple late decision-level fusion \cite{lee2020cbb} and co-attention \cite{bi2020multimodal}. 

Bidirectional Encoder Representations from Transformers (BERT) is a state-of-the-art open-source language model\cite{adhikari2019docbert}. BERT Models are fine-tuned for specific goals. Sun et al. \cite{sun2020finetune} demonstrated improved text classification performance by using various strategies, such as layer selection, layer-wise learning rate, etc. BERT has been used in a variety of ways to improve language processing tasks. Variations of BERT, trained on large French corpora, include FlauBERT \cite{Hang} and CamemBERT \cite{Martin}. 

In this study, we will fine-tune the pre-trained French language models for each textual modality separately and the SE -ResNeXt-50 \cite{Hu_2018_CVPR} for image products to extract features for each modality. The extracted features are hierarchically fused using different fusion strategies to make the prediction.

\section{Hierarchical Fusion Model}
\label{prop}
In this section, we describe our proposed model architecture for classifying product types. 

The core of our method is to project the unimodal representation sequences generated by the transformers $P$, $T_f, T_c$ and $D_f, D_c$ into a common representation space and then learn a classifier in this common space:

\begin{equation}
y = f (Z(P, Z(Z(T_f,D_f), Z(T_c,D_c) ) )),
\label{multimodal_equation}
\end{equation}
where $Z$ is a fusion function that combines the extracted features, and $f$ is a neural network classifier that maps the features of the resulting joint space to a class prediction $y$. $P$ is the reduced image representation in SE -ResNeXt-50 \cite{Hu_2018_CVPR}, $T_f, T_c, D_f, D_c$ are the representation sequences extracted from the titles and the descriptions by FlauBERT \cite{Hang} and CamemBERT \cite{Martin}, respectively.

The proposed hierarchical model uses state-of-the-art pre-trained architectures in the base models. In developing the fusion model, we have tried to keep it as simple as possible, since a simple architecture can provide good performance with less complexity and cost. The fusion architecture is a simple trainable module.

\subsection{Text Encoding Models}

We fine-tuned the two French-language models, FlauBERT and CamemBERT, for encoding textual information of the product title and description into the textual sequence representation. FlauBERT was trained with unsupervised masked language modeling (MLM, similar to BERT), while CamemBERT was pre-trained on French text based on Facebook's RoBERTa model.

Given a textual product title sentence $T = ( t_1, t_2, \dots t_n)$, FlauBERT and CamemBERT give  embedding sequences $T_f = (t_{f1}, t_{f2}, \dots, t_{fn})$,  and $T_c = (t_{c1}, t_{c2}, \dots , t_{cn})$, while for a given textual product description $D = (d_1, d2,\dots, dm)$, FlauBERT and CamemBERT produce  $D_f = (d_{f1}, d_{f2}, \dots, dfm)$  $D_c = (D_{c1}, D_{c2}, \dots, D_{cm})$ embedding sequences, respectively.

\subsection{Image Encoding Model}

For the purpose of obtaining representations of product images, we use the SE-ResNeXt-50 \cite{Hu_2018_CVPR} model. The last fully-connected (FC) layer of the pre-trained model has been replaced, and the model has been fine-tuned for our categorization task (that will be removed for the feature extraction). Based on the work of \cite{jimaging7080157,Wang2017,Tashu}, an input image $I$ is re-sized to $224\times224$ and divided into $16 \times 16$ regions. A raw image vector, a.k.a. a regional feature representation, is then produced by applying the SE-ResNeXt-50 model to each region $I_i (i = 1, 2, \dots, 256)$. The final image feature vector ($P$) is then obtained by averaging all regional image vectors.

\begin{equation}
    P = \frac{\sum \limits_{i=1}^{N_r} \text{ResNeXt}(I_i)} {N_r} 
\end{equation}

\noindent where ResNeXt$(I_i)$ is the raw image vector extracted via SE-ResNeXt-50 from region $i$, $N_r$ ($256$ in this work) is the number of regions. The final visual representation $P_{\text{full}}$ is the average of all regional image vectors.

\subsection{Multi-modal Joint Representation Learning}


$P$ represents the visual features extracted by running the input images through the SE-ResNeXt-50 model. We used a 1D convolution layer with max-pooling  ($P\in\mathbb{R}^d$) to fit the dimension of the image representation to the textual representations. Similarly, textual features extracted by passing titles to CamemBERT and FlauBERT models are represented as  $T\_c, T\_f \in (\mathbb{R}^{d})$ and descriptions extracted by passing descriptions through CamemBERT and FlauBERT models are represented as $D_f, D_c \in ( \mathbb{R}^{d})$. To combine the text and image features, we implemented the following fusing functions ($Z$):  
\begin{itemize}

\item \textbf{Addition Fusion:}  The sum of input representations is defined as the addition operation. Given two inputs  ($x_1, x_2$) of any modalities with the same dimension $d$, the addition operation yields the output \(X_{add} = x_1 + x_2 \), where \(X_{add} \in \mathbb{R}^{d}\).

\item \textbf{Concatenation Fusion:} The concatenation operation is defined as the concatenation of representations across a dimension. In terms of the base models: 
$X_{con} = < x_1 , x_2 > $ , where $X_{con} \in \mathbb{R}^{d\times 2}$ and $ <>$ is the concatenation operation.

\item \textbf{Average Fusion:}  
The mean of the input representations is used to define the average operation.\\
\(X_{avg}(x_1,x_2) = mean( x_{{1}} , x_{{2}} )\), where \(X_{avg} \in \mathbb{R}^{d}\). 

\end{itemize}

 \begin{figure*}[ht]
    \centering
    \includegraphics[width = 1 \textwidth ]{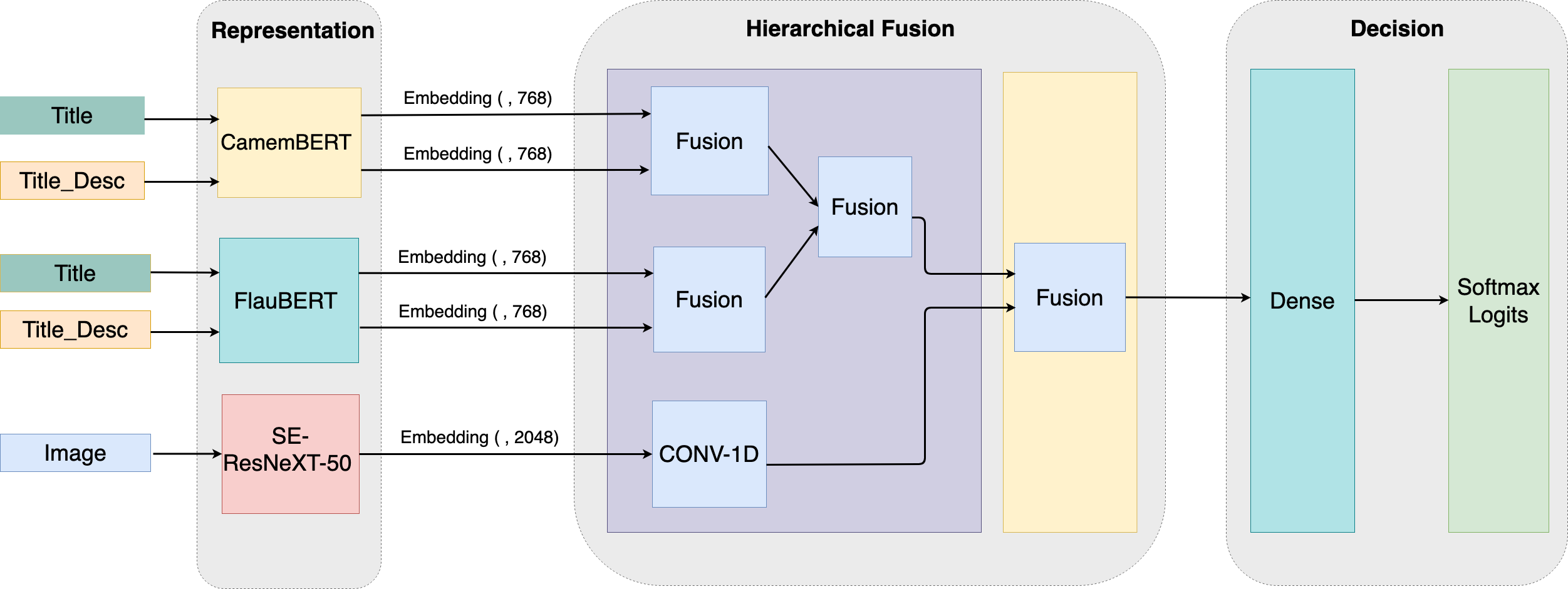}
    \caption{Multi-Modal Representation: Hierarchical Fusion }
    \label{proposed}
\end{figure*}

\subsection{Architecture}

Our suggested strategy involves training a fully connected neural network model that uses the hierarchically fused embedding vectors generated by the three unimodal networks to construct a new probability vector over the target categories. Our research aims to discover the most effective strategies to combine the representations discovered by several distinct models. The proposed architecture shown in Figure \ref{proposed} consists of the following layers: 

\begin{itemize}

\item \textbf {Input layer:}  The input layer receives and preprocesses visual and textual modalities . We feed the textual base models with the product titles and descriptions individually, resulting in four different textual embeddings, and split the input images into regions before passing them to SE-ResReXt-50.\\

\item\textbf {Representation layer:}  Using fine-tuned unimodal networks, the Representation layer extracts feature vectors from preprocessed input data. It also combines the SE-ResReXt-50 outputs by region.\\

\item \textbf{Intermediate fusion layer:} The embeddings of the text modalities (product title and description) learned from each language model (CamemBERT, FlauBERT) were first fused using one of our fusion methods in the Intermediate fusion layer, then the fused results from CamemBERT and FlauBERT were further fused to obtain the final vector representation for text modalities.\\

\item \textbf{Feed forward layer}. Finally, the fused vector is passed through a three-layer fully connected neural network to produce a classification result. The classifications are generated using the softmax activation function.

\end{itemize}

When training the fusion model, the uni-modal network layers are frozen, and the refined product description and title representations are combined to produce a single refined representation, which, along with the reduced image representation, serves as input to our fully connected neural network. The fusion layers' operations produced various variants of the proposed hierarchical architecture.

\section{Experimental Settings}
\label{experimental}
\subsection{Dataset}\label{dataset}

We used a dataset provided by the Rakuten Institute of Technology for the Rakuten Multi-modal Product Data Classification and Retrieval Challenge to test the effectiveness of our proposed model. This task's data set contains 99K products, nearly 84K of which were included in the training dataset. There are 27 different product type codes in the text and image dataset, totaling 55K unique products. The four parent domains that group the entire product catalog are Children, Household, Entertainment, and Books. The text data is in French and consists of the title and description of the product.

The description field has a mean length of 35 words and a maximum length of 2068 words, while the title field has a mean length of 11 words and a maximum length of 56 words. The product images are all square, with white or black borders and a $500 \times 500$ pixel resolution. Each product is associated with an image. We preprocessed the data and performed the following steps: 

\begin{itemize}
    \item Text Pre-processing: We tokenize the product title and description using the NLTK \footnote{http://www.nltk.org} tokenizer, lowercase the text and remove HTML tags. 
    
     \item Product images: To fine tune the pretrained mage models were augmented by randomly rotating, flipping, and extracting random crops from the original images. 
     
    \item Product images were resized to a uniform size of 224 × 224 and normalized per channel using ImageNet mean and standard deviation.

\end{itemize}

\begin{figure}[h]
    \centering
    \includegraphics[width=0.5 \textwidth ]{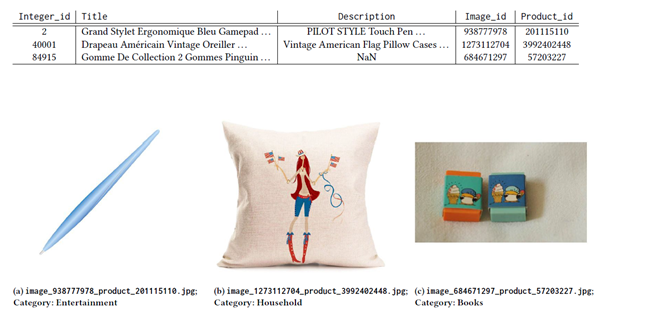}
    \caption{Example of the multimodal information in the training set}
    \label{Example_products}
\end{figure}

\subsection{Implementation Details}

PyTorch was used to create all of our models. We employed the 138M parameter $FlauBERT_{base}$ architecture with encoded vocabulary, as well as the 110M parameter $CamemBERT_{base}$ architecture. With a batch size of 32 and a sequence length of 256 for 10 epochs, the FlauBERT and CamemBERT models were fine-tuned With a batch size of 32 for 20 epochs, the SE -ResNeXt-50 model was fine-tuned as well.

The pre-trained weights of the trained uni-modal base models were used to fine-tune the learning module for the multimodal representation. The uni-modal network layers were frozen, and we used feature fusion methods to train our suggested multimodal design, followed by a fully connected. network with a Softmax output layer to perform product classification. As a loss function, we employed the categorical cross-entropy minimization objective, the AdamW optimizer with a learning rate value of $2 \times 10^{-5}$ for text models, and the Adam optimizer with $10^{-3}$ for image models.

\section{Results and Discussion} 
\label{res}

Multi-modal models were created with the goal of combining the representations given by uni-modal models to produce more accurate, reliable, and robust predictions. Several versions have been developed, with varying levels of overall performance. In the following sections, we present our findings. We split the original training set and utilized it for training and testing because the labels for the test set were not available.\\

We employed two alternative splits to train and test our proposed methods in order to assess the generalization power of our model. We used 90\%  for training and 10\% for testing in the first phase. 10\% from the training was also used as a validation set. We used 80\% for training and 20\% for testing in the second phase. A validation set made up of 10\% of the 80 percent training was also utilized. The 90/10 split was utilized to create a test set that was around the same size as the challenge.

\subsection{90/10, Train/test Split}

In the first experiment, we used a 90/10 split between training and testing, using 90\% for training  and 10\% for testing the model. Table \ref{Baseline_UNI_9010tabble} shows the performance of each uni-modal model, namely CamemBERT-title, CamemBERT-description, FlauBERT-title, FlauBERT-description, and SE -ResNeXt individually.
Table\ref{Baseline9010tabble} shows the multimodal hierarchical fusion models on the test set using the concatenation/average fusion method in the first fusion layer (fusion of text embeddings) and different combination methods in the last fusion layer.\\

\begin{table}[htbp]
\caption{Macro F1 scores and Accuracy of uni-modal models on test Set - 90/10 Split}
\centering
\begin{tabular}{ |c|c|c| } 
\hline
\textbf{Model}  & \textbf{Accuracy} & \textbf{F1 Score} \\
\hline
\multicolumn{3}{|c|}{\textbf{Validation Set}}\\
\hline

 CamemBERT\_{description}   &  86.98\%  & 84.83\% \\
 
 CamemBERT\_{title}          &90.19\%    &87.38\% \\

FlauBERT\_{description}     &  84.43\%  & 84.43\%  \\
            
FlauBERT\_{title}     &92.31\%   &   90.32\%  \\   

SE-ResNeXt-50  &59.25\%  &54.30 \% \\ 
\hline
\end{tabular}
\label{Baseline_UNI_9010tabble}
\end{table}

\begin{table}[htbp]
\caption{ Hierarchical Fusion Results for various groupings of models on test Set - 90/10 Split}
\centering
\begin{tabular}{ |c|c|c| } 
\hline
\textbf{Model}  & \textbf{Accuracy} & \textbf{F1 Score} \\
\hline
\multicolumn{3}{|c|}{\textbf{Concatenation Based }}\\      
\hline

Concatenation Fusion &  92.80\%  & 92.47\% \\ 

Addition Fusion   &  92.75\%  &  92.45\%  \\ 

Average Fusion  &  92.86\%  &  92.47\%   \\ 

\hline


\multicolumn{3}{|c|}{\textbf{Average Based }}\\      
\hline
\textbf { Average Fusion}   & \textbf{93.20\%}  & \textbf{92.67\% }  \\ 
  \hline            

\end{tabular}
\label{Baseline9010tabble}
\end{table}

In the proposed hierarchical fusion architecture, it is noticeable that the model using the average combination in both fusion layers performs well. 

\subsection{Introducing More Layers}

After running different experiments, the model using the average fusion method in both fusion layers showed the best performance compared to other models trained with a 90/10 split of the dataset. We started  incrementally experimenting  with Decision Layer by adding additional layers to further improve the performance of this model as follows: 

\begin{itemize} 
    \item Adding a dropout layer after the last fusion layer.
    \item Stacking a dropout layer followed by a fully connected (FC) layer and a non-linearity before the final projection layer.

\end{itemize}

Table \ref{more} presents the performance of the basic proposed architecture and the modified architecture having stacked dropout out, FC and non-linearity after the final fusion layer.
\begin{table}[htbp]
\caption{Average Fusion performance with Different Architectures - 90/10 Split}
\centering
\begin{tabular}{ |c|c|c| } 
\hline
 \textbf{Model}  & \textbf{Accuracy} & \textbf{F1 Score} \\
\hline
\multicolumn{3}{|c|}{\textbf{Validation Set}}\\
\hline

\textbf { Basic Model  }    &\textbf{93.20\%}      &\textbf{92.67\%}  \\ 
 
With Dropout    &92.49\%    &90.00\% \\ 

With More Layers    &92.42\%    &91.11\% \\ 

\hline  

\end{tabular}

\label{more}
\end{table}

As we can see in Table \ref{more}, the performance of the models decreased when we added more additional layers. Therefore, the modified model with more layers did not seem to improve the results, while the simplest architecture (Figure \ref{proposed}) showed better overall performance and reduced the training time, giving room for further experiments. \\

It is  noticeable in the simplest hierarchical architecture that the model that used average combination in both fusion layers has a good performance on the test set. However, to obtain a robust model, further investigation is needed to verify that the model is not overfitted. To address the issue, we split the dataset in a way that allows us to obtain more test set, which is the second split strategy mentioned in section \ref{res}. 

\subsection{80/20, Train/Test Split}

This spilt was performed to see how the models performs on less training set and a bit more test set.  Table \ref{proposed8020_UNI} and table \ref{proposed8020_multi} shows the performance of uni-modal models and the proposed hierarchical model variations in terms of accuracy and F1 Score on the  test set of the given dataset split, using Concatenation/Average fusion method  in  the  first  fusion  layer  (textual  embeddings' fusion)  and  different  combination  methods  in  the  final fusion layer.

\begin{table}[htbp]
\centering
\caption{  Macro F1 scores and Accuracy of uni-modal models on test Set - 80/20 Split}
\begin{tabular}{ |c|c|c| } 
\hline
\textbf{Model}  & \textbf{Accuracy} & \textbf{F1 Score} \\
\hline
\multicolumn{3}{|c|}{\textbf{test Set}}\\
\hline

CamemBERT\_{description}                &86.40\%    &83.13\% \\ 

CamemBERT\_{title}                      &89.64\%    &84.01\% \\

FlauBERT\_{description}                 &85.81\%   &81.08\% \\ 

FlauBERT\_{title}                       &88.98\%   &83.10\%  \\    
SE-ResNeX-50                           &57.20\%   &52.32\%\\ 
\hline  
\end{tabular}
\label{proposed8020_UNI}
\end{table}

\begin{table}[htbp]
\centering
\caption{  Hierarchical Fusion Models Performance on Test Sets- 80/20 Split}
\begin{tabular}{ |c|c|c| } 
\hline
\textbf{Model}  & \textbf{Accuracy} & \textbf{F1 Score} \\
\hline

\multicolumn{3}{|c|}{\textbf{Concatenation Based }}\\      
\hline
Addition Fusion                       & 91.97\%  &89.39\% \\
             
{Concatenation Fusion } &{92.36\%}  &   {90.03\%} \\ 
  
\textbf{Average Fusion} & \textbf{92.86\%}  &   \textbf{92.46\% } \\ 

\hline


\multicolumn{3}{|c|}{\textbf{Average Based }}\\      
\hline
{ Average Fusion}   & {92.25\%}  & {89.94\% }  \\ 
\hline

\end{tabular}
\label{proposed8020_multi}
\end{table}

We had the best performance when we used concatenation in the first and then averaging in the second fusion layer in the 80/20 split as shown in table \ref{proposed8020_multi}, but full average fusion outperformed the concatenation based models in the validation and test sets as evidenced in table \ref{Baseline9010tabble}. The best model has achieved an overall F1 Score of \(92.67\)\%, which indicates that merging the two modalities boosts the performance compared to uni-modal networks.

\section{Conclusion}
\label{con}

In this work, multiple Deep Learning based pre-trained and fine-tuned models were used with different fusion strategies with the goal of producing a simple, easily trainable and good quality model for multimodal e-commerce product classification.  We used a dataset provided by Rakuten Institute of Technology for the Rakuten Multi-modal Product Data Classification and Retrieval Challenge that contains 27 different product type codes and a total of 55K unique products.

The best performance was obtained using the average fusion operation in all fusion layers, where the model reached a F1-Score of \(92.67\)\% on the test set given 90/10 dataset split. When it came to  80/20 dataset split, the best achieved model was concatenation-based in fusing textual features, and average-based fusion in the final fusion layer with F1-Score of  \(92.46\)\%. Our experimental results in general showed that fine-tuning pre-trained models for feature vector representation with simple hierarchical fusion approaches can yield good results in multimodal e-commerce product classification.

\section*{{Acknowledgements}}

This research is supported by the \'UNKP-21-4 New National Excellence Program of the Ministry for Innovation and Technology from the source of the National Research, Development and Innovation Fund. 

Supported by the Telekom Innovation Laboratories (T-Labs), the Research and Development unit of Deutsche Telekom.

Project no. ED\_18-1-2019-0030 (Application domain specific, highly reliable IT solutions subprogramme) has been implemented with the support provided from the National Research, Development and Innovation Fund of Hungary, financed under the Thematic Excellence Programme funding scheme.

\bibliographystyle{plain}
\bibliography{bibliography.bib}

\end{document}